\documentclass[sigconf]{acmart}

\AtBeginDocument{%
  }

\setcopyright{acmlicensed}
\copyrightyear{2024}
\acmYear{2024}
\acmDOI{XXXXXXX.XXXXXXX}

\acmConference[KDD WorkShop]{Make sure to enter the correct
  conference title from your rights confirmation email}{August 25-26, 2024}{Barcelona, Spain}
\acmISBN{978-1-4503-XXXX-X/18/06}

\usepackage{algorithm}  
\usepackage{algorithmicx}  
\usepackage{algpseudocode} 
\usepackage{diagbox}
\usepackage{booktabs}
\usepackage{color}
\usepackage{graphicx}
\usepackage{subcaption}
\usepackage{multirow}
\usepackage{amsmath}
\usepackage{setspace}
\usepackage{enumitem}

\usepackage{amssymb}
\usepackage{amsmath}

\DeclareMathOperator*{\argmin}{arg\,min}

\usepackage{color}
\usepackage{xcolor}

\begin{document}

\title{Improving Fairness in Graph Neural Networks via Counterfactual Debiasing}

\author{Zengyi Wo}
\email{wozengyi1999@tju.edu.cn}
\orcid{0009-0007-6939-0042}
\affiliation{%
  \institution{Tianjin University}
  \city{Tianjin}
  \country{China}
}

\author{Chang Liu}
\affiliation{%
  \institution{Tianjin University}
  \city{Tianjin}
  \country{China}
}

\author{Yumeng Wang}
\affiliation{%
  \institution{Tianjin University}
  \city{Tianjin}
  \country{China}
}

\author{Minglai Shao}
\authornote{Corresponding author}
\email{shaoml@tju.edu.cn }
\affiliation{%
  \institution{Tianjin University}
  \city{Tianjin}
  \country{China}
}

\author{Wenjun Wang}
\authornote{Corresponding author}
\email{wjwang@tju.edu.cn}
\affiliation{%
  \institution{Tianjin University}
  \city{Tianjin}
  \country{China}
}

\begin{abstract}
Graph Neural Networks (GNNs) have been successful in modeling graph-structured data. However, similar to other machine learning models, GNNs can exhibit bias in predictions based on attributes like race and gender. 
Moreover, bias in GNNs can be exacerbated by the graph structure and message-passing mechanisms. Recent cutting-edge methods propose mitigating bias by filtering out sensitive information from input or representations, like edge dropping or feature masking. Yet, we argue that such strategies may unintentionally eliminate non-sensitive features, leading to a compromised balance between predictive accuracy and fairness. To tackle this challenge, we present a novel approach utilizing counterfactual data augmentation for bias mitigation. This method involves creating diverse neighborhoods using counterfactuals before message passing, facilitating unbiased node representations learning from the augmented graph. Subsequently, an adversarial discriminator is employed to diminish bias in predictions by conventional GNN classifiers. Our proposed technique, Fair-ICD, ensures the fairness of GNNs under moderate conditions. Experiments on standard datasets using three GNN backbones demonstrate that Fair-ICD notably enhances fairness metrics while preserving high predictive performance.
\end{abstract}



\maketitle

\section{Introduction}
Graph-structured data, e.g., citation networks, has become increasingly pervasive in the real world. In recent years, Graph Neural Networks (GNNs) have garnered significant attention for their remarkable achievements in modeling such data. The representations learned through GNNs are pivotal in a diverse array of downstream tasks (e.g., node classification
\cite{xiao2022graph,wo2024graph}and link prediction\cite{zhang2018link,rossi2021knowledge}) and real-world applications (e.g., drug discovery \cite{gaudelet2021utilizing}and anomaly detection\cite{ma2021comprehensive,akoglu2015graph}), generating substantial societal impact. 

Despite powerful representational capabilities, recent studies\cite{dai2021say, fairview} have found that the predictions of GNNs lack consideration for fairness, potentially leading to discriminatory and biased decisions. The application of such unfair predictions to high-stakes decision-making tasks, e.g., crime prediction\cite{jin2020addressing} and credit assessment\cite{yeh2009comparisons}, could engender significant ethical and moral issues. The biases in the predictions of GNNs can be attributed to their propensity to inherit and exacerbate the biases embedded in historical data. \textit{Bias from feature data}. The statistical correlation between node features and sensitive information consequently leads to node embeddings implicitly encoding sensitive attribute information. \textit{Bias from the graph structure}. According to the homophily effect, nodes with the same sensitive attributes are more likely to be connected compared to nodes with different sensitive attributes. \textit{Aggregation Mechanisms for GNNs}. The aggregation mechanism of GNNs smoothes the representations of neighboring nodes, further causing the convergence of representations of nodes with the same sensitive attributes, ultimately exacerbating the correlation between prediction derived from representations and sensitive attributes.

Over the last few years, there have been numerous studies exploring the enhancement of fairness in GNNs. Generally, the core idea of these methods lies in completely discarding the information related to sensitive attributes (e.g., edge dropping\cite{agarwal2021towards}, feature masking\cite{fairview}), with the expectation that the prediction outcomes of GNNs become independent of sensitive attributes to fulfill fairness requirements. This method, based on the concept of entirely discarding sensitive attribute-related information, often struggles to fully separate information related to sensitive attributes but irrelevant to downstream tasks and valid information (information genuinely relevant to downstream tasks). Consequently, it inevitably leads to the loss of information pertinent to downstream tasks, resulting in a sub-optimal balance between informativeness (The ability of the representation to predict labels) and fairness. 

To tackle the concern mentioned above, we suggest adopting a fresh approach to increase the diversity of node neighborhoods by utilizing a bias offsetting technique and addressing bias in GNN classification with the help of an adversarial discriminator. Our unique strategy focuses on counterfactual data augmentation as a means to combat bias. This approach involves generating varied neighborhoods with counterfactuals prior to message propagation to acquire unbiased node representations from the augmented graph. Furthermore, we utilize adversarial training of the discriminator to minimize biased predictions in standard GNN classifiers. Our methodology is termed Fair-ICD.

Our contributions are as follows:(1) We propose a novel paradigm based on \textit{bias offsetting} for learning fair Graph Neural Networks (GNNs), which leverages data augmentation to enhance the heterogeneity of neighborhoods to mitigate sensitive attributes. (2) Specifically, we achieve fair representation learning in three GNN backbones through counterfactual reasoning combined with adversarial discriminators. (3) Experimental results demonstrate that compared to recent state-of-the-art methods, the proposed Fair-ICD can achieve a better trade-off between predictive performance and fairness. 

\section{Related Work}
\subsection{Fairness in Graphs}
Most machine learning models lack considerations for fairness, potentially resulting in biased and unfair decisions. Graph mining algorithms, exemplified by GNNs, also suffer from the same issue. For example, job recommendation models may preferentially allocate opportunities to applicants of a specific gender group, even when candidates from different gender groups possess similar qualifications relevant to job performance. Fairness can be categorized into several common types: \textit{group fairness}\cite{hardt2016equality}, where algorithms should neither discriminate against nor favor any sensitive subgroup; \textit{individual fairness}\cite{dwork2012fairness}, where similar individuals receive similar treatment; and \textit{counterfactual fairness}\cite{kusner2017counterfactual}, where fair decisions are independent of sensitive attribute values. 

EDITS\cite{dong2022edits} is a pre-processing method that proposes a model-agnostic framework to provide any GNN with a bias-reduced attributed network as input. NIFTY\cite{agarwal2021towards} generates two augmented graphs by slightly perturbing node attributes and edges, as well as flipping sensitive attribute values to create counterfactuals of the original nodes. 
Graphair\cite{ling2023learning} proposes an automated augmentation model to generate augmented graphs, utilizing both adversarial learning and contrastive learning to achieve fairness and informativeness. FairVGNN\cite{fairview} considers the impact of sensitive information leakage after feature propagation. It automatically learns fair views by identifying and masking sensitive-related channels and adaptively clamping weights. FairGNN\cite{dai2021say} ensures that the representations generated by GNNs do not leak sensitive information through adversarial debiasing, while also adding a covariance constraint.
As observed, \textit{edge perturbation}, \textit{feature masking}, and \textit{adversarial debiasing} are common and effective methods for enhancing fairness. Nonetheless, the core idea of most existing approaches is to completely discard sensitive attribute-related information. Due to the complex entanglement of information, ensuring that no relevant information for downstream tasks is lost in this process is challenging. 

\subsection{Counterfactual Augmentation in Graphs}
In the task of exploring fairness, a counterfactual\cite{guo2023counterfactual, agarwal2021towards, ma2022learning, guo2023towards} node refers to a version of the original node with different sensitive attributes. Specifically, it is a node with a different sensitive attribute value but the same label (similar features) as the original node. There are three methods for obtaining counterfactual nodes: generation, modeling, and searching. 

\textbf{Generation.} Generation-based methods typically create counterfactuals by flipping the sensitive attributes. NIFTY\cite{agarwal2021towards} generates counterfactual augmented graphs by perturbing the sensitive attributes of nodes. Although generation methods are simple, they rely on the unrealistic assumption that sensitive attributes have no causal influence on other attributes and the graph structure, resulting in counterfactuals that do not truly exist. Simply flipping sensitive attributes may disrupt the original semantic information, thus affecting the final performance. 
\textbf{Modeling.} Model-based methods reconstruct the dependency between the graph structure and sensitive attributes after flipping the sensitive attributes. They modify the graph's structure or feature matrix to obtain counterfactual augmented graphs. GEAR\cite{ma2022learning} generates two types of counterfactual augmented graphs through GraphVAE after perturbing the sensitive attribute values of the node itself and neighbors. Despite considering the dependency between graph structure and sensitive attributes, the counterfactuals obtained by this method are still non-realistic. 
\textbf{Searching.} Search-based methods emphasize finding suitable counterfactuals within training data. The counterfactuals obtained through this approach genuinely exist, and the information they contain is more realistic and reliable. CAF\cite{guo2023towards} utilizes labels and sensitive attributes as guidance to search for potential counterfactuals in the representation space.

\section{PRELIMINARY}
\subsection{Background}
\subsubsection{Graph}
 Let $\mathcal{G} = \left ( \mathcal{V} ,\mathcal{E}, \mathrm{A}, \mathrm{X}  \right ) $ denote an attributed graph, comprised of a set of $N$ nodes $\mathcal{V}$ and a set of edges $\mathcal{E}$. Here, $\mathrm{X} \in \mathbb{R}^{N\times D}$ denotes the node attribute matrix and  $\mathrm{A} \in \mathbb{R}^{N\times N}$ is the adjacency matrix where $\mathrm{A}_{ij}=1$ if edge $e_{ij} \in \mathcal{E}$ between node $v_i$ and $v_j$ exists, otherwise $\mathrm{A}_{ij}=0$. Each node $v_i$ has a sensitive attribute $s_i \in \left \{ 0, 1 \right \} $. 
 \subsubsection{Graph Neural Networks (GNNs)}
 The neighbor set of node $v_i$ is denoted as $\mathcal{N}(i) = \{ v_j \mid (v_i, v_j) \in \mathcal{E} \}$. Graph Neural Networks (GNNs) aim to create node representations by iteratively passing information from neighboring nodes. In a general message-passing scheme, the representation of node $v_i$ is updated as follows:
\begin{equation}
z_i^{(k)} = \text{UPDATE}(z_i^{(k-1)}, \text{AGGREGATE}(z_j^{(k-1)} \mid j \in \mathcal{N}(i))),
\end{equation}
where the functions $\text{AGGREGATE}(\cdot)$ and $\text{UPDATE}(\cdot)$ are trainable and responsible for neighbor aggregation and representation update, respectively.

In Graph Convolutional Network (\textbf{GCN})\cite{kipf2016semi}, the aggregation process incorporates a mean aggregator, while the update function involves a linear transformation followed by a non-linear activation. 
\textbf{GIN}\cite{xu2018powerful} employs a sum aggregator to capture complete neighborhood information and a multi-layer perceptron (MLP) for the update function. On the other hand, \textbf{GraphSAGE}\cite{hamilton2017inductive} utilizes a variety of aggregation techniques, such as mean, LSTM, or pooling aggregators, and integrates the node's own characteristics with the aggregated neighborhood features prior to the application of the update function.
\begin{figure}[h!]
    \centering
    \includegraphics[width=0.8\linewidth]{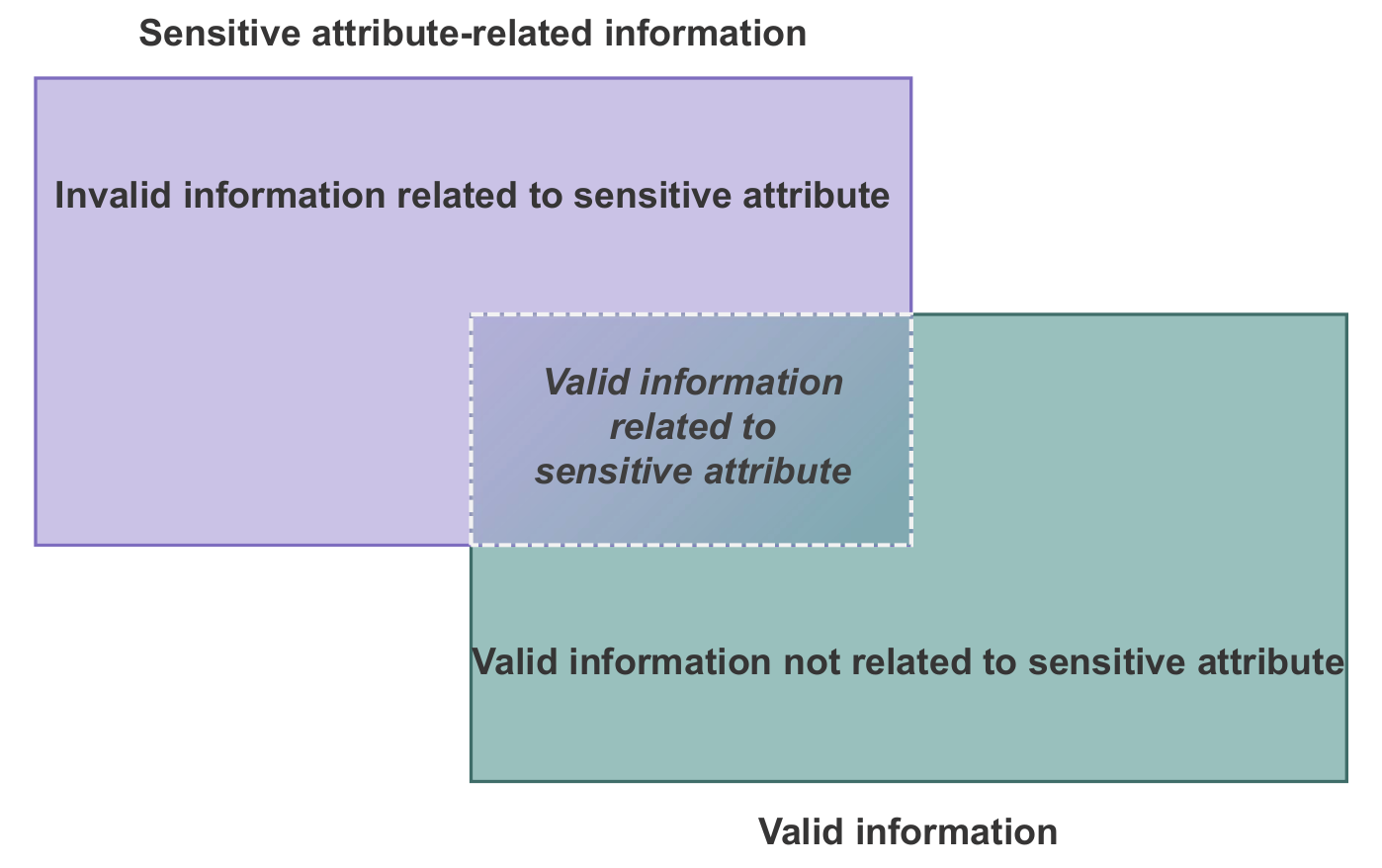}
    \caption{Complex entanglement between sensitive attribute-related information and valid information.}
    \label{figure1}
\end{figure}

\subsubsection{Fairness Assessment in Node Classification.}
\textbf{Demographic Parity}\cite{dwork2012fairness}. Intuitively, $\mathbf{DP}$ is a common metric used to measure the disparity in acceptance rates between two sensitive subgroups:
\begin{equation}
\mathrm{DP} = \left | P\left ( \hat{Y} = 1 | S = 0 \right ) - P\left ( \hat{Y} = 1 | S = 1 \right ) \right |.  
\end{equation}

\noindent\textbf{Equality of Opportunity}\cite{fairview,dai2021say,dong2022edits}. This metric recognizes that in many scenarios, sensitive features may be relevant to the target task, requiring positive predictions to be independent of the sensitive features of individuals with positive true labels:
\begin{equation}
\mathrm{EO} = \left | P\left ( \hat{Y} = 1 | Y = 1, S = 0 \right ) - P\left ( \hat{Y} = 1 | Y=1, S = 1 \right ) \right |.  
\end{equation}

\subsection{Fairness in GNNs}
\label{3.2}
\begin{table}[]
    \centering
    \caption{The performance of three data augmentation strategies concerning informativeness and fairness on the Pokec-n dataset. Arrows ($\nearrow$, $\searrow$) indicate performance that underperforms or exceeds vanilla. Arrows ($\uparrow$, $\downarrow$) indicate the direction of better performance.}
    \label{table_1}
    \begin{tabular}{c|cc}
        \hline
        \hline    
        Strategies &  ACC ($\uparrow$) & DP ($\downarrow$) \\
        \hline
        vanilla &  68.55±0.51 & 3.75±0.94 \\
        Edge-dropping &  67.24±0.49 ($\searrow$) & 1.22±0.94 ($\nearrow$)\\
        Feature-masking &  66.10±1.45 ($\searrow$) & 1.69±0.79 ($\nearrow$) \\
        \textbf{Bias-offsetting} & \textbf{69.06±0.6} ($\nearrow$)  & \textbf{0.67±0.39} ($\nearrow$)\\
        \hline
        \hline
    \end{tabular}  
\vspace{-0.3cm}
\end{table}
\begin{table}[]
    \centering
    \caption{The bias evaluation of the original graph and counterfactual augmented graph on the Pokec-n dataset. Let $\mathcal{G}$ denote the Original graph and $\mathcal{G'}$ denote the Counterfactual augmented graph. Arrows ($\uparrow$, $\downarrow$) indicate the direction of better performance.}
    \label{table_2}
    \begin{tabular}{c|cc}
        \hline
        \hline    
         & $\mathcal{G}$ & $\mathcal{G'}$ \\
        \hline
        Avg. degree & 16.53 & 16.53 \\
        Avg. heterogeneous degree ($\uparrow$) &  0.73 & 8.13 \\
        Nodes w/o heterogeneous neighbors ($\downarrow$) &  46134 & 8183 \\
        \hline
        \hline
    \end{tabular} 
\vspace{-0.3cm}
\end{table}

\subsubsection{Motivations}
 We investigated the performance of three data augmentation strategies concerning informativeness and fairness: Edge-dropping, Feature-masking, and Bias-offsetting, as shown in Table \ref{table_1}. 
\begin{itemize}[leftmargin=*]
    \item \textbf{Edge-dropping}, a common strategy for data augmentation and processing, involves removing some edges from the graph;
    \item \textbf{Feature-masking}, which is a data augmentation technique that involves concealing certain node features, compelling the model to depend on the remaining information at its disposal;
    \item \textbf{Bias-offsetting}, which encourages central nodes to focus more on aggregating neighbor nodes with different sensitive attributes in an attempt to offset biased and invalid information related to sensitive attributes and mitigate bias.
\end{itemize}
\noindent\textbf{Results} 
Table \ref{table_1} shows the performance of GNN after different data augmentations: 
\begin{itemize}[leftmargin=*]
\item in terms of \textbf{ACC}: $\text{\textbf{Bias-offsetting }}>\text{\textbf{Vanilla}}>\text{\textbf{Edge-dropping}}>\text{\textbf{Feature-masking } }$; 

\item in terms of \textbf{DP}: $\text{\textbf{Bias-offsetting }}<\text{\textbf{Vanilla}}<\text{\textbf{Edge-dropping}}<\text{\textbf{Feature-masking } }$.
\end{itemize}
Table \ref{table_2} shows enhancing the Average heterogeneous degree, decreasing Nodes without heterogeneous neighbors.
 We calculated the average heterogeneous degree to measure the degree of bias in the graph. Note that in this context, heterogeneity refers to nodes having different sensitive attribute values. 
\begin{equation}
\text{Avg. heterogeneous degree}  = \frac{\sum_{v_i\in \mathcal{V}}^{}| \{v_j \in \mathcal{N}\left ( i \right ) |s_i \ne s_j\}|}{|\mathcal{V}|},
\end{equation}
 The lower the average heterogeneous degree, the higher the degree of connection between nodes with the same sensitive attribute values, leading to a higher correlation between representations and sensitive attributes, thus increasing the risk of bias.
 
\noindent\textbf{Remarks}
Based on our observations, we have determined that the edge-dropping and feature-masking techniques do not consistently improve both informativeness and fairness when compared to standard methods. This challenge stems from the intricate interplay between sensitive attribute data and pertinent information, as depicted in Figure \ref{figure1}. While edge-dropping and feature-masking strive to eliminate sensitive attribute data entirely, this process also discards relevant information crucial for subsequent tasks, thus creating an inadequate equilibrium between informativeness and fairness. Conversely, our bias-offsetting approach effectively enhances both dimensions concurrently.

\subsubsection{Task}
 \textit{Informativeness.} In this research, we investigate the semi-supervised node classification task. Specifically, our goal is to develop a model capable of predicting the label $\hat{y} \in \mathcal{Y}$ for each node in the unlabeled node set $\mathcal{V}^U = \mathcal{V} \setminus \mathcal{V}^L$, utilizing the graph $\mathcal{G}$, node features $\mathrm{X}$, and the labeled node set $\mathcal{V}^L \subseteq \mathcal{V}$. A common model architecture involves combining a GNN encoder with a classifier. The classification performance assessment entails comparing the actual labels $Y$ with the predicted labels $\hat{Y}$. \textit{Fairness.} The primary aim concerning fairness is to minimize the correlation between predicted labels $\hat{Y}$ and sensitive attributes $S$ while preserving classification accuracy as much as possible, equivalent to reducing the correlation between node representations and sensitive attributes. 

\section{METHODOLOGY}
\noindent\textbf{Overview} This section provides a detailed explanation of our proposed method, named Fair-ICD, which aims to enhance the fairness of GNN learning representations through counterfactual data augmentation. It consists of two modules: the Unbiased Representation Learning Module and the Adversarial Debiasing Module. In the Unbiased Representation Learning Module, we use counterfactual methods to enhance the heterogeneity of the original graph neighbors, enabling the learning of unbiased representation patterns for nodes from the augmented graph. In the Adversarial Debiasing Module, adversarial training is employed to reduce the bias in predictions made by the GNN classifier. Now, we will elaborate on each design.

\subsection{Unbiased Representation Learning}
To achieve a fair (unbiased) representation of node features successfully, we propose a counterfactual enhancement method. By investigating the impartiality of nodes in the aggregation process, we aim to enhance the influence of node neighborhood heterogeneity, enabling the manipulation of augmented graph generation through counterfactual interventions.

\subsubsection{Counterfactual Data Augmentation}
Our objective is to mitigate bias stemming from sensitive attribute information by ensuring that counterfactuals possess contrasting sensitive attributes. While a straightforward approach involves flipping the sensitive attribute of a given sample to create a counterfactual, this method is generally ineffective. To overcome this limitation, as detailed in section \ref{3.2}, we introduce a novel data augmentation technique that emphasizes the distribution of node neighborhoods within the original graph. When dealing with nodes having distinct sensitive attributes, we maintain the edges; conversely, when nodes share the same sensitive attribute, we identify counterfactual nodes for the neighborhood node and establish counterfactual relationships among these identified nodes:
\begin{align}
    (v_a, v_b)=\argmin_{v_a, v_b\in\mathcal{V}}\  \big[  \|\mathbf{x}_a-\mathbf{x}_b\|_2^2|s_a\neq s_b\big],
\end{align}
where $v_b$ represents the counterfactual node of $v_a$, which can also be expressed as $v_a^c = v_b$.

The counterfactually augmented graph $\mathcal{G}^{\prime}$ is defined as follows:
\begin{align}
    \mathcal{G}^{\prime} = \begin{cases}
    A_{ij}, & \text{if } s_i \neq s_j \\
    A_{ik}, & \text{if } s_i = s_j \text{ and } v_j^c = v_k
    \end{cases},
\end{align}

In practical implementation, we initially assess similarity by calculating the Euclidean distance between node features. Subsequently, we identify the Top-k nodes that are most similar to each node and then pinpoint nodes within this set that possess distinct sensitive information. By applying the counterfactual processing described above, we enhance the graph such that the neighborhoods surrounding central nodes exhibit heterogeneity.

\subsubsection{Fairness Neighbor Capturing}
Due to the existence of nodes where high-quality counterfactual nodes cannot be found, we use a simple $MLP_{\theta}$ to learn the pattern of enhanced high-quality heterogeneous neighborhood aggregation:
\begin{equation}
\mathcal{L}_\text{unbias}= \argmin\mathbb{E}_{i, j\sim \mathcal{V}} ||MLP_{\theta}(x_i)-\text{AGG}(x_i,A^{\prime}_{ij})||,
\end{equation}
in this context, $\text{AGG}(\cdot,\cdot)$ denotes mean aggregation, while $A^{\prime}_{ij}$ signifies the adjacency matrix of the augmented graph.

Our goal is to obtain unbiased and fair representations for nodes in the original graph. Since the original features contain a lot of useful semantic information, we concatenate the original features with unbiased features and input them into a traditional GNN encoder $f_G$ for message-passing aggregation:                                                    
\begin{equation}
\tilde{\mathbf{X}}^k=\mathbf{X}^k+\operatorname{MLP}_\theta^k(\mathbf{X}^k),\\
\mathbf{X}^{k+1}=f_G(\tilde{\mathbf{X}}^k).
\end{equation}

\subsection{Adversarial Debiasing}
The GNN classifier $f_G$ can make biased predictions because the learned representation of $f_G$ exhibits bias due to the node features, graph structure, and aggregation mechanism of GNN. One approach to ensure fairness in $f_G$ is to eliminate bias in the final layer representation. To achieve this, we introduce an adversarial discriminator that assists the GNN classifier in debiasing. We use binary cross-entropy loss to provide constraints for the discriminator:
\begin{equation}
\min \mathcal{L}_d=-\frac1{|\mathcal{V}|}\sum_{v\in\mathcal{V}}[s_v\log\hat{s}_v+(1-s_v)\log{(1-\hat{s}_v)}].
\end{equation}

\section{Experiments}
\subsection{Experimental Settings}
\subsubsection{Datasets}
Following the approaches proposed in \cite{dong2022edits}, we evaluate our work and baseline methods on the real-world benchmark datasets: Pokec-n\cite{takac2012data}. These datasets have been extensively used in previous studies on graph fairness learning, and cover a diverse range. We provide the dataset statistics in Table \ref{Dataset statistics}.

\begin{table}[]
    \centering
    \caption{Dataset statistics.}
    
    \begin{tabular}{c|cc}
        \hline
        \hline    
        Dataset&  Pokec-n  \\
        \hline
        Nodes&  66569  \\
        Edges&  729129\\
        Features&  266 \\
        Lable& Working filed \\
        sensitive attribute& Region\\
        \hline
        \hline
    \end{tabular}
    
    \label{Dataset statistics}
    \vspace{-0.5cm}
\end{table}

\subsubsection{Baselines}
We will compare our methods with the latest cutting-edge techniques for enhancing fairness.
\begin{itemize}[leftmargin=*]
    \item \textbf{Vanilla}: The encoder in our approach is built upon three widely used graph neural networks (GNNs): GCN \cite{kipf2016semi}, GIN \cite{xu2018powerful}, and GraphSAGE \cite{hamilton2017inductive}.
    \item \textbf{FairGNN}\cite{dai2021say}: This method focuses on mitigating bias through adversarial training.
    \item \textbf{EDITS}\cite{dong2022edits}: An approach that reduces discrimination among various sensitive groups by modifying the graph structure and node attributes.
    \item \textbf{NIFTY}\cite{agarwal2021towards}: A technique that combines feature perturbation and edge dropping to ensure fairness by enhancing similarity between augmented and counterfactual graphs.
    \item \textbf{FairVGNN}\cite{fairview}: A framework designed to prevent leakage of sensitive attributes by concealing correlated channels and adjusting weights adaptively.
\end{itemize}
\subsubsection{Evaluation Protocol}
We evaluate the performance of downstream classification tasks using F1 score and accuracy metrics. When assessing group fairness, we consider DP  and EO based on previous research. It is essential to understand that lower values of DP and EO indicate a higher level of fairness in a model.
\subsubsection{Model Hyperparameters and Implementation Details}
We utilize a Multilayer Perceptron (MLP) in our method to predict the features of neighboring entities. Our focus lies on counterfactual data augmentation, specifically targeting the Top-k values of \{3, 5, 10, 25\}. For the MLP optimization, we utilize the Adam optimizer and adjust the learning rate within \{0.1, 0.01, 0.001\}. The hyper-parameter coefficient in our approach is fine-tuned within the range of [0, 10]. The GNN encoders are configured following the settings presented in \cite{fairview}. Results are presented as the mean and standard deviation from five runs with diverse random seeds. All experiments are conducted on a single GPU unit featuring a GeForce GTX 3090 with 24 GB of memory.

\subsection{Experimental Results}
\begin{table}[]
    \centering
    \caption{Node classification Experimental results(mean with standard deviation) comparison on datasets Pokec-n.}
    \resizebox{1.0\linewidth}{!}{
    \begin{tabular}{c|c|cccc}
        \hline
        \hline    
        \multirow{2}{*}{Method} & \multirow{2}{*}{Model} & \multicolumn{4}{c}{Pokec-n} \\
        & & F1 & Acc & DP & EO \\
        \hline
        \multirow{6}{*}{GCN} & Vanilla & \textbf{67.74±0.41} & \underline{68.55±0.51} & 3.75±0.94 & 2.93±1.15 \\
        & FairGNN & 65.62±2.03 & 67.36±2.06 & 3.29±2.95 &2.46±2.64 \\
        & EDITS & OOM & OOM & OOM & OOM \\
        & NIFTY & 64.02±1.26 & 67.24±0.49 &\underline{1.22±0.94} & 2.79±1.24 \\
        & FairVGNN & 64.85±1.17 &66.10±1.45 & 1.69±0.79 & \underline{1.78±0.70} \\
        &\textbf{Fair-ICD} & \underline{67.55±0.41} & \textbf{69.06±0.60 }&\textbf{ 0.67±0.39 }& \textbf{0.82±0.66} \\
        \hline
        \hline
        \multirow{6}{*}{GIN} & Vanilla &67.87±0.70 & \underline{69.25±1.75} & 3.71±1.20 & 2.55±1.52 \\
        & FairGNN &64.73±1.86 &67.10±3.25& 3.82±2.44 &3.62±2.78 \\
        & EDITS & OOM & OOM & OOM & OOM  \\
        & NIFTY & 61.82±3.25&66.37±1.51& 3.84±1.05 &3.24±1.60\\
        & FairVGNN & \underline{68.01±1.08} &68.37±0.97&\underline{1.88±0.99} &\underline{1.24±1.06} \\
        &\textbf{Fair-ICD} & \textbf{68.06±0.97} &\textbf{69.67±0.61} & \textbf{1.36±0.39}&\textbf{0.99±0.49 }\\
        \hline
        \hline
        \multirow{6}{*}{GraphSAGE} &Vanilla & 67.15±0.88&\underline{69.03±0.77} &3.09±1.29 &2.21±1.60 \\
        & FairGNN & 65.75±1.89 &67.03±2.61& 2.97±1.28 &2.06±3.02 \\
        & EDITS & OOM & OOM & OOM & OOM  \\
        & NIFTY & 61.70±1.47 &68.48±1.11 &3.84±1.05& 3.90±2.18 \\
        & FairVGNN & \underline{67.40±1.20}& 68.50±0.71 &\textbf{1.12±0.98} &\underline{1.13±1.02} \\
        &\textbf{Fair-ICD} & \textbf{68.35±0.89}& \textbf{69.33±0.53}& \underline{1.22±0.46} &\textbf{1.08±1.12} \\
        \hline
        \hline
    \end{tabular}
    }
    \label{experimental results}
\end{table}

We showcase the findings of Fair-ICD to illustrate that our approach, centered on counterfactual bias mitigation, can attain a more favorable balance compared to the state-of-the-art (SOTA) method. Displayed in Table \ref{experimental results}, Fair-ICD demonstrates superior overall classification performance and fairness across various GNN encoders. Concerning fairness, Fair-ICD diminishes both DP and EO in comparison to the top-performing baseline. Furthermore, in numerous instances, Fair-ICD can surpass standard encoders in accuracy metrics, aligning well with our rationale and model architecture.

\section{conclusion}
In this paper, we propose an innovative strategy based on counterfactual data augmentation to achieve bias offsetting, where a heterogeneous neighborhood is constructed using counterfactuals before message passing, allowing unbiased representations of nodes to be learned from the augmented graph. Subsequently, adversarial training is employed to reduce the bias in predictions of traditional GNN classifiers. We name our approach Fair-ICD, which ensures fairness of GNN under mild conditions. Experimental results on benchmark datasets with three different GNN backbones show that Fair-ICD significantly improves fairness metrics while maintaining high prediction accuracy.

\begin{acks}
This work is supported by the National Natural Science Foundation of China (No.62272338).
\end{acks}

\bibliographystyle{ACM-Reference-Format}
\bibliography{sample-base}

\end{document}